\pdfoutput=1

\documentclass[11pt]{article}

\usepackage[final]{acl}

\usepackage{booktabs}
\usepackage{array}
\usepackage{url}
\usepackage{longtable}
\usepackage{pdflscape}
\usepackage{geometry}
\usepackage{multirow}

\usepackage{times}
\usepackage{latexsym}

\usepackage[T1]{fontenc}

\usepackage[utf8]{inputenc}

\usepackage{microtype}

\usepackage{inconsolata}

\usepackage{graphicx}
\usepackage{amsmath}

\title{When Babies Teach Babies: Can student knowledge sharing outperform Teacher-Guided Distillation on small datasets?}

\author{Srikrishna Iyer \\
  Artificial Intelligence - Data Analytics Strategic Technology Center, \\
  ST Engineering IHQ Ltd., Singapore \\
  \texttt{srikrishna.rameshiyer@stengg.com}}

\begin{document}
\maketitle
\begin{abstract}
We present our submission\footnote{\url{https://huggingface.co/AI-DA-STC/RoBERTa_WML_distill-Babylm-10M-2024}}\footnote{\url{https://github.com/AI-DA-STC/generative-ai-research-babylm}} to the BabyLM challenge, aiming to push the boundaries of data-efficient language model pretraining. Our method builds upon deep mutual learning, introducing a student model search for diverse initialization. We address the limitation of treating students equally by formulating weighted mutual learning as a bi-level optimization problem. The inner loop learns compact students through online distillation, while the outer loop optimizes weights for better knowledge distillation from diverse students. This dynamic weighting strategy eliminates the need for a teacher model, reducing computational requirements. Our evaluations show that teacher-less methods can match or surpass teacher-supervised approaches.  
\end{abstract}

\section{Introduction}

The substantial computational and memory requirements of large language models pose significant challenges for deployment on intelligent edge systems, where resources are often constrained. As the demand for real-time processing and low-latency responses increases in edge computing environments, the need for lightweight and memory-efficient models becomes critical. Recent research, notably the Chinchilla paper (\citet{hoffmann_training_2024}), demonstrated that a 70B parameter model trained on 1.4 trillion tokens outperformed larger models with less data, highlighting the intricate balance between model size and training data. This massive data requirement—equivalent to over 10,000 times the words a 13-year-old encounters—is becoming a significant bottleneck. To address these challenges, several techniques have emerged such as network pruning (\citet{han_learning_2015-1}), quantization (\citet{courbariaux_binaryconnect_2015}), neural architecture search \citet{ren_comprehensive_2021} and Knowledge distillation (\citet{hinton_distilling_2015},\citet{vedaldi_local_2020},\citet{wang_head_2022})

In response to these challenges, the BabyLM challenge invites researchers to explore the limits of data-efficient language model pretraining \cite{choshen_call_nodate}. Participants are constrained to training their models on limited text corpora: 10M and 100M word text-only tracks and a newly introduced multimodal track containing 50M words of paired text-image data, and 50M words text-only data.

Our paper describes our submission to the 10M and 100M text-only tracks. It builds upon the approach of weighted mutual learning \citet{zhang_weighted_nodate} while introducing key modifications to enhance generalizability. Our methodology focuses on distilling a RoBERTa-base model (125M parameters) to less than half its size while maintaining performance. Our main contributions include : 
\begin{itemize}
    \item We use Bayesian optimization to select model architectures of student models by varying hidden layers, attention heads, and hidden sizes.
    \item Instead of the traditional teacher-student distillation, we explore weighted mutual learning through a bi-level optimization process : 
    (a) The inner loop minimizes a combined loss to train individual student models, consisting of a supervised learning loss and a KL divergence loss that aligns each student's class posterior with others'. 
    (b) Instead of treating each student model equally, we introduce an outer loop to optimize student importance weights by minimizing the ensemble loss.
\end{itemize}  

This approach generally performed better than both conventional supervised learning and traditional distillation from a larger pretrained teacher. Notably, our weighted mutual learning strategy can improve performance even among several large networks compared to independent learning, challenging the conventional understanding that distillation requires a larger, more powerful teacher.

\begin{figure*}[t]
  \centering
  \includegraphics[width=\textwidth]{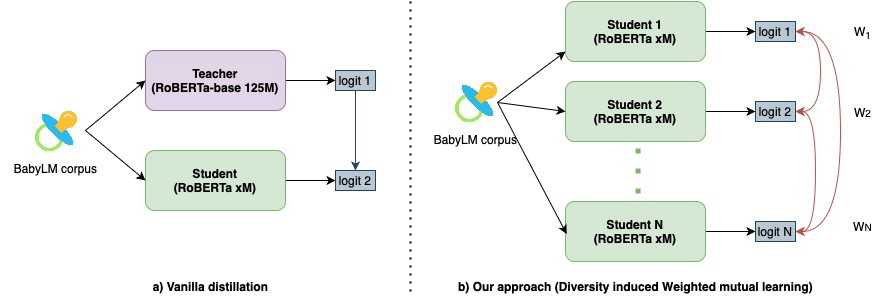}
  \caption{Overview of the difference between Vanilla knowledge distillation and our approach, Diversity induced weighted mutual learning (DWML). (a) \citet{hinton_distilling_2015} is the popular knowledge distillation method, where the student network (RoBERTa-xM) can only learn from a trained teacher network (RoBERTa-base-125M). Here xM refers to a student model of x million parameters. (b) is the Diversity Induced Weight Mutual Learning (DWML)
  framework where each student model is initialised with parameter counts = $N/2, N/3 .. N/(p+1)$ using Bayesian optimisation search. Rather than averaging the knowledge from students, DWML leverages bi-level optimization to estimate the relative importance of each student (e.g., weight $\omega_i$ for student $i$).}
  \label{fig:enter-label}
\end{figure*}

\section{Related Work}

The vanilla distillation \citet{hinton_distilling_2015} method consists of two stages, firstly train a large teacher model, followed by transfer of soft logits to a smaller student model. Also known as Offline distillation, it keeps the teacher fixed, only allowing a one-way knowledge transfer. To reduce memory consumption of training a large teacher model, \citet{zhang_deep_2018} proposed an online distillation framework called mutual learning where a group of student (or student) models were trained simultaneously. Although, online distillation eliminated the teacher model, similar networks in online distillation may prevent the students from learning knowledge from the students \citet{zhang_weighted_nodate}. 
Recent approaches have attempted to induce diversity in online distillation to improve overall performance. \citet{chen_online_2020} proposed inducing data diversity by training student models with varying image augmentations. However, this method relies heavily on data augmentations, which can be unpredictable in real-world deployment scenarios. \citet{du_agree_2020} introduced an adaptive ensemble knowledge distillation method using multiple diverse teacher models to train a student model. While this approach shows promise, it requires maintaining several teacher models, leading to increased memory usage and computational overhead. The reported accuracy improvements are also relatively modest, typically ranging from 0.5\% to 1\% across benchmarks. Our approach closely resembles to that of \citet{zhang_weighted_nodate}. They present a diversity induced weight mutual learning approach for distillation. They introduce diversity by assigning varying pruning ratios to different student models. Although this method reduces memory consumption, the manual assignment of pruning ratios may not generalize well across different architectures and tasks. The reported performance gains are limited, with improvements of less than 0.5\% on most benchmarks. As shown by \citet{liu_learning_2017}, while pruning induces sparsity within networks and can reduce computational complexity (measured in FLOPs), the relationship between pruning percentage and actual model size reduction is not always linear. Moreover, in \citet{zhang_weighted_nodate}, we observe a performance drop when pruning beyond 30\%, indicating a trade-off between model compression and accuracy. 

\section{Diversity Induced Weighted Mutual Learning}

\subsection{Diversifying student models}
In our approach to create diverse student models for the Diversity Induced Weight Mutual Learning (DWML) framework, we employ Bayesian optimization to efficiently search for optimal architectural configurations. Given a teacher model with N parameters, we aim to generate p student models, where the i-th student model targets approximately $N_i$ parameters, defined as:
\begin{equation}
N_i = \frac{N}{i+1}, \quad i \in {1, 2, ..., p}
\end{equation}
This optimization problem can be formally defined as finding, for each student i, an architecture $a_i$ from the set of all possible RoBERTa architectures A that minimizes $||params(a_i) - N_i||$, where $params(a_i)$ represents the parameter count of architecture $a_i$.
We chose Bayesian optimization for this task due to its efficiency in exploring high-dimensional spaces with relatively few function evaluations, making it less computationally expensive compared to alternative methods such as grid search or random search \cite{kandasamy_neural_2018}. Our implementation utilizes the BayesianOptimization library \cite{fernando_bayesian_opt_2014}, with a search space encompassing the number of layers, number of attention heads, and embedding dimension. The objective function calculates the difference between the actual parameter count of a given architecture and the target parameter count, with a constraint ensuring the embedding dimension is divisible by the number of attention heads.

\subsection{Weighted Mutual Learning using Bi-level optimisation}

Building upon the work of \cite{zhang_weighted_nodate}, we introduce a modified approach to Weighted Mutual Learning using bi-level optimization. Our method replaces the pruning-based initialization with Bayesian optimization for student model selection.

The overall loss function for training M peer models is defined as:
\begin{equation}
\begin{split}
\text{loss} = (1 - \alpha)\sum_{i=1}^M \omega_i L_{CE}(z_i, Y) \\
+ \alpha\sum_{i=1}^M\sum_{j=1}^M \omega_j KL(z_i, z_j)
\end{split}
\label{loss}
\end{equation}

where $\omega_i$ indicates the importance of the i-th student model, $\alpha$ balances the supervision from labels and peers, $L_{CE}$ is the cross-entropy loss, and $KL$ is the Kullback-Leibler divergence. $\omega_j$ is the importance of every other student model except the i-th one. Both $z_i$ and $z_j$ are model logits. We formulate the weighted mutual learning as a bi-level optimization problem. The inner loop optimizes the network parameters $\theta$ using the loss in equation \ref{loss}. As shown in the paper, the gradient for the outer loop optimization, also known as the hypergradient, is calculated as:

\begin{equation}
    g_{\omega_i} = \nabla_{\omega_i} L_2 = \frac{\partial L_2}{\partial\omega_i} - \gamma \frac{\partial L_2}{\partial\theta}\frac{\partial L_a}{\partial\theta}^T
\label{gradient}
\end{equation}

where $L_a = (1 - \alpha)L_{CE}(z_i, Y) + \alpha \sum_{j=1}^M KL(z_j, z_i)$ is the ensemble loss. Since $\omega$ is a probability simplex that $\sum_{i=1}^M \omega_i = 1$, we use the mirror descent to update $\omega$ [3, 5]. Algorithm 1 outlines our weighted mutual learning for online distillation. To be more specific, we first run several steps of gradient descent based on the loss function in \ref{loss} to update model parameters $\theta$ with a fixed $\omega$. Then we calculate the gradient of $\omega_i$ based on \ref{gradient}, and run one step of mirror descent to update $\omega_i$:

\begin{equation}
    \omega_i^{k+1} = \frac{\omega_i^k \exp\{-\eta\nabla_{\omega_i^{k+1}} L_2\}}{\sum_{i=1}^M \omega_i^k \exp\{-\eta\nabla_{\omega_i^{k+1}} L_2\}}
\label{mirror}
\end{equation}

where $\eta$ is the step size with annealing, and $\omega_i^k$ is the importance of the i-th peer in the k-th step.

\begin{table}[h]
\centering
\begin{tabular}{lc}
\hline
\textbf{Algorithm 1:} Diversity Induced \\
Weighted Mutual Learning (DWML)\\
\hline
\textbf{Input:} Dataset $\{(x_n,y_n)\}_n^N$; Number of peers \\ \quad\quad\quad M; Teacher model size N \\
1: Initialize parameter space, $\theta = \{layers \in N^+,$\\ $heads \in N^+, dim \in N^+\}$\\
3: \textbf{for} each student model $i = 1$ to $M$ \textbf{do} \\
4:$\quad \theta_i^* = \arg\min_{\theta} \left|\text{size}\{S_i(\theta)\} - \frac{M}{i+1}\right|$ \\ \quad\quad $\text{ Subject to:  layers | heads}$\\
5:\quad Initialize peer model $i$ with parameters $\theta_i^*$ \\
6: \textbf{end for} \\
7: Initialize peer weights $\omega^0$ \\
8: \textbf{for} $k = 1$ to $K$ \textbf{do} \\
9:\quad  With peer importance $\omega^k$, run T steps of \\ \quad\quad AdamW to update model parameters $\theta$ using \\ \quad\quad Eq. \ref{loss} \\
10:\quad Calculate gradient for $\omega^k$ based on Eq. \ref{gradient} \\
11:\quad Update $\omega^k$ to $\omega^{k+1}$ using mirror descent \\ \quad\quad with Eq. \ref{mirror} \\
12: \textbf{end for} \\
\textbf{Output:} M models with outputs ${z_1, ..., z_M}$ \\ and weights for peers $\omega$ \\
\hline
\end{tabular}
\end{table}

\section{Training}

\subsection{RoBERTa-base}
Our models are based on RoBERTa-base \cite{liu_roberta_2019}. This model has shown reasonably good performance on small text corpus. We use the raw RoBERTa-base as a baseline in the evaluations. We use it as our teacher model to distill student models using knowledge distillation (KD) and the teacher supervised version of weighted deep mutual learning (KD\_DWML). Details about the hyperparameters found from the search are shown in \ref{tab:pretraining-hyperparameters}. The models were pre-trained (and finetuned for GLUE, SuperGLUE tasks) using 1 Nvidia H100 GPU with 80GB VRAM.  

\subsection{Dataset}
We pretrain all our language models on the 10M and 100M datasets of the BabyLM challenge from 2023 \cite{warstadt_findings_2023}. We adopt the same preprocessing pipeline from \cite{samuel_trained_2023} for standardizing the text corpus. The detailed breakdown of the datasets are shown in \ref{tab:data}. The reason why we select the datasets from 2023 is that it appears to be similar to the dataset released for the 2024 challenge \cite{choshen_call_nodate}. The only difference is the exclusion of the QCRI Educational Domain (QED) Corpus and higher proportion of CHILDES from 4.21M to 29M. This was done because the QED was of poor quality. However, we believe that the 2023 dataset gives us an opportunity to explore how distilled models perform when trained datasets that closely represent real world textual data that is unavoidably noisy.  

\section{Results}

This section provides the results of the empirical evaluation of DWML. First, we compare our method to baselines, then we compare our method with other distillation methods and then we perform an ablation study of different DWML variations.

\subsection{BabyLM Challenge evaluation}

\begin{table}[htbp]
\centering
\small
\begin{tabular}{p{2cm}cccc}
\hline
\multicolumn{5}{c}{\textbf{Text-only 10M Dataset}} \\
\hline
Model & BLiMP & Supp. & EWoK & GLUE \\
\hline
BabyLlama & \textbf{69.8} & 59.5 & 50.7 & \textbf{63.3} \\
LTG-BERT & 60.6 & \textbf{60.8} & 48.9 & 60.3 \\
RoBERTa-base & 49.6 & 48.9 & \textbf{51.6} & 42.5 \\
RoBERTa-DWML & 51.6 & 52.3 & 50.3 & 43.1 \\
\hline
\multicolumn{5}{c}{\textbf{Text-only 100M Dataset}} \\
\hline
Model & BLiMP & Supp. & EWoK & GLUE \\
\hline
BabyLlama & \textbf{73.1} & 60.6 & 52.1 & \textbf{69.0} \\
LTG-BERT & 69.2 & \textbf{66.5} & 51.9 & 68.4 \\
RoBERTa-base & 49.8 & 46.8 & 50.25 & 43.4\\
RoBERTa-DWML & 52.1 & 48.4 & 51.6 & 44.0 \\
\hline
\end{tabular}
\caption{Results for the BabyLM challenge evaluation datasets. We compare our submitted model (RoBERTa-DWML) to the base model (RoBERTa-base) and the baselines given by the organizers of the challenge on the 10M and 100M datasets.}
\label{tab:babylm-results}
\end{table}

We use the BabyLM evaluation pipeline to assess our models. This pipeline measures syntactic understanding through the Benchmark of Linguistic Minimal Pairs (BLiMP \& BLiMP supplemental,\citet{warstadt_blimp_2020}). It evaluates general knowledge using the Elements of World Knowledge (EWoK, \citet{ivanova_elements_2024}) benchmark. For overall natural language understanding, it uses GLUE (\citet{wang_glue_2018}) and SuperGLUE (\citet{wang_superglue_2019}. If applicable, we divide the training set into a train-development split and report the mean statistics over multiple runs on the hidden validation split.The detailed scores are shown in section \ref{detailed results}.

\paragraph{BLiMP}
Our RoBERTa-DWML demonstrates consistent improvements over RoBERTa-base across both dataset sizes. On the 10M dataset, DWML achieves 51.6\% compared to RoBERTa-base's 49.6\%, showing a 2\% improvement. This gain is maintained in the 100M dataset, where DWML scores 52.1\% versus RoBERTa-base's 49.8\%. While these improvements are modest, they demonstrate that our teacher-less approach can enhance syntactic understanding with minimal computational overhead. It's worth noting that BabyLlama's multi-teacher distillation approach \cite{timiryasov-tastet-2023-baby} significantly outperforms all models (73.1\% on 100M), though this comes at the cost of substantial computational requirements in maintaining and training with multiple teacher models (GPT-2 and LLaMA), which may not be practical for resource-constrained applications. 
\paragraph{BLiMP Supplemental}
The supplemental BLiMP results further validate the effectiveness of our DWML approach. For the 10M dataset, RoBERTa-DWML (52.3\%) outperforms RoBERTa-base (48.9\%) by a margin of 3.4\%. In the 100M setting, we observe a similar trend with DWML (48.4\%) showing improvement over the base model (46.8\%). These consistent gains come with minimal additional computational cost over the base model. While BabyLlama achieves substantially higher performance (60.6\% on 100M), this improvement requires significant computational resources for managing multiple teacher models during training and inference, a trade-off not examined in their original work.
\paragraph{EWoK}
On the world knowledge tasks, RoBERTa-DWML maintains competitive performance relative to RoBERTa-base. In the 10M dataset, DWML (50.3\%) performs slightly below the base model (51.6\%), while in the 100M dataset, DWML (51.6\%) shows improvement over the base model (50.25\%). These results demonstrate the capability of our lightweight approach in preserving world knowledge. While BabyLlama leads with 52.1\% on the 100M dataset through its multi-teacher architecture, the relatively small performance gap (0.5\%) raises questions about whether the significant computational overhead of maintaining multiple teacher models is justified for world knowledge tasks in resource-constrained environments.

\paragraph{GLUE}
All the models were fine-tuned on the GLUE and SuperGLUE datasets and then evaluated on their linguistic performance.
On the GLUE benchmark, RoBERTa-DWML shows marginal improvements over RoBERTa-base across both dataset sizes. For the 10M dataset, DWML achieves 43.1\% compared to RoBERTa-base's 42.5\%, representing a modest 0.6\% gain. This pattern continues in the 100M setting, where DWML (44.0\%) slightly outperforms the base model (43.4\%). These results suggest that our teacher-less approach maintains general language understanding capabilities

\begin{table*}[htbp]
\centering
\small
\begin{tabular}{lcccccc}
\hline
\multicolumn{7}{c}{\textbf{BLiMP Filtered}} \\
\hline
Method & Teacher & Peer 1 $(60M)$ & Peer 2 $(42M)$ & Peer 3 $(34M)$ & Peer 4 $(28M)$ & Best$\uparrow$  \\
\hline
RoBERTa-base-125M & - & - & - & - & - & 49.62 \\
SD & No & 51.73 & 50.04 & 50.31 & 51.18 & \textbf{51.73} \\
KD & Yes & 46.47 & 47.25 & 47.09 & 47.65 & 47.65 \\
DML & No & 47.01 & 47.77 & 47.21 & 47.16 & 47.44 \\
KD\_DWML (Ours) & Yes & 47.05 & 47.28 & 47.47 & 46.66 & 47.47 \\
DWML (Ours) & No & 50.45 & 51.58 & 51.46 & 50.63 & 51.58 \\ \\
\hline
\multicolumn{7}{c}{\textbf{BLiMP Supplement}} \\
\hline
Method & Teacher & Peer 1 $(60M)$ & Peer 2 $(42M)$ & Peer 3 $(34M)$ & Peer 4 $(28M)$ & Best$\uparrow$ \\
\hline
RoBERTa-base-125M & - & - & - & - & - & 48.9 \\
SD & No & 53.03 & 54.78 & 49.63 & 56.53 & \textbf{56.53}\\
KD & Yes & 53.73 & 52.64 & 52.58 & 55.82 & 55.82 \\
DML & No & 44.74 & 45.14 & 45.19 & 44.96 & 45.19 \\
KD\_DWML (Ours) & Yes & 52.21 & 53.09 & 53.34 & 53.65 & 53.65\\
DWML (Ours) & No & 52.25 & 48.99 & 48.43 & 47.99 & 52.25\\ \\
\hline
\multicolumn{7}{c}{\textbf{EWoK Filtered}} \\
\hline
Method & Teacher & Peer 1 $(60M)$ & Peer 2 $(42M)$ & Peer 3 $(34M)$ & Peer 4 $(28M)$ & Best$\uparrow$ \\
\hline
RoBERTa-base-125M & - & - & - & - & - & 51.6 \\
SD & No & 48.4 & 49.38 & 50.36 & 49.19 & 50.36 \\
KD & Yes & 50.12 & 50.3 & 51.56 & 50.42 & 51.56 \\
DML & No & 50.05 & 50.12 & 50.06 & 48.82 & 50.12 \\
KD\_DWML (Ours) & Yes & 55.44 & 40.36 & 50.75 & 49.83 & \textbf{55.44} \\
DWML (Ours) & No & 49.98 & 49.84 & 49.08 & 50.29 & 50.29 \\ \\
\hline
\end{tabular}
\caption{BLiMP Filtered, BLiMP Supplement, and EWoK scores for Text-only 10M dataset, comparing different distillation methods. Best accuracy scores (higher is better) are shown.}
\label{tab:combined-results}
\end{table*}

\subsection{Comparison with Other Distillation Methods}

To evaluate the effectiveness of our proposed distillation method, in Table \ref{tab:combined-results} we compare its performance against other distillation techniques using accuracy scores. Our framework is compared to Self-Distillation (SD, \citet{zhang2019teacherimproveperformanceconvolutional}), a method that allows a small-sized student model to distill knowledge within its network. Knowledge distillation (KD,\citet{hinton_distilling_2015}) is the vanilla distillation framework that uses a student network to approximate the output logits of a pretrained teacher network. Deep mutual learning (DML,\citet{zhang_deep_2018}) an ensemble of students learn collaboratively (without a teacher) and teach each other. The main difference between DML and our diversity induced weight mutual learning (DWML) framework is the usage of dynamically learned student weights using a bi-level optimization objective.Knowledge distillation based diversity induced weight mutual learning (KD\_DWML) is the teacher-supervised version of DWML. The GPU utilization and training times are shown in Table \ref{tab:training_utilization} and Figure \ref{fig:gpu_utilization}. They clearly show a trade-off between training times(mins) and GPU Utilization(\%). While our approach DWML had the lowest GPU utilization among all, the training time was reported the highest. 
\paragraph{BLiMP Filtered}
On the BLiMP Filtered dataset, teacher-less methods demonstrate superior performance, with SD and DWML achieving 51.73\% and 51.58\% respectively, significantly outperforming their teacher-supervised counterparts KD (47.65\%) and KD\_DWML (47.47\%). Among all approaches, our DWML framework shows strong performance, ranking second only to SD with a marginal difference of 0.15\%. Notably, DWML substantially outperforms traditional KD by 3.93\% and DML by 4.14\%, validating the effectiveness of our dynamic weighting strategy in the absence of teacher supervision. Compared to the RoBERTa-base baseline (49.62\%), both teacher-less methods show clear improvements, with DWML achieving a 1.96\% gain, suggesting that peer learning alone can enhance syntactic understanding.
\paragraph{BLiMP Supplement}
The BLiMP Supplement results further reinforce the advantage of teacher-less methods, with SD achieving the highest score of 56.53\%. Our DWML method (52.25\%) outperforms DML (45.19\%) by a substantial margin of 7.06\%, though it falls behind SD. While KD (55.82\%) and KD\_DWML (53.65\%) show competitive performance, the superior performance of SD demonstrates that teacher supervision isn't necessary for strong syntactic understanding. All distillation methods except DML surpass the RoBERTa-base baseline (48.9\%) by a significant margin, with our DWML showing a 3.35\% improvement, further validating the effectiveness of peer learning for syntactic tasks.
\paragraph{EWoK Filtered}
On the EWoK Filtered dataset, we observe a unique pattern where KD\_DWML achieves the highest performance (55.44\%), though teacher-less methods still show strong consistency, with SD, DML, and DWML achieving 50.36\%, 50.12\%, and 50.29\% respectively. Interestingly, teacher-less methods perform slightly below the baseline, with a performance gap of up to 1.24\%. This deviation from the pattern observed in BLiMP datasets suggests that world knowledge tasks may benefit more from teacher guidance, which could explain why KD\_DWML achieved the best performance with a substantial 3.84\% improvement over the baseline. This finding indicates that while peer learning is effective for syntactic tasks, world knowledge acquisition might require the structured guidance that teacher supervision provides.

\subsection{Ablation studies}

We compare the following modifications to the original DWML architecture :
\begin{enumerate}
    \item \textbf{Varying number of students} : The effect of using different number of student networks during training.
    \item \textbf{Varying $\alpha$ ratio between label and peer supervision} : The effect of using different $\alpha$ in equation \ref{loss} that balances KL divergence and cross-entropy loss. 
    \item \textbf{Effect of dynamic student weights} : Determining if learning peer weights during training affect model performance (average accuracy \%)
    \item \textbf{Effect of model size} : Determining if model sizes affected model performance (average accuracy \%)
\end{enumerate} 

\begin{figure*}[t]
    \centering
    \includegraphics[width=\textwidth]{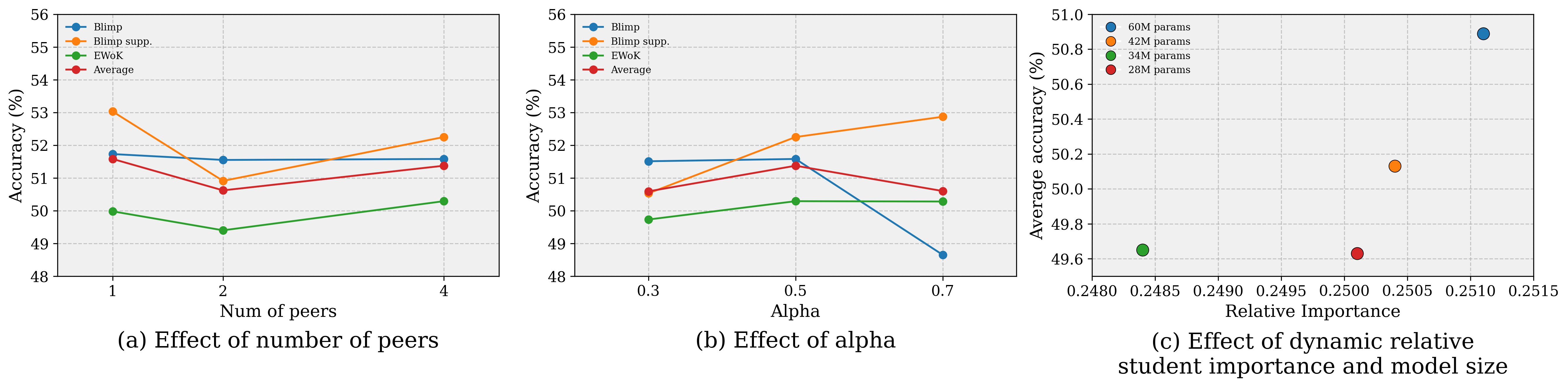}
    \caption{Performance comparison across different experimental settings for 10M dataset: (left) varying number of peers, showing how model performance changes with different peer counts; (middle) impact of alpha parameter in the loss function on model accuracy; (right) relationship between relative importance and accuracy for different model sizes.}
    \label{fig:ablation_study}
\end{figure*}

\paragraph{Effect of Varying number of student models}
Figure \ref{fig:ablation_study}(a) illustrates the impact of increasing the number of peer networks in our DWML framework. Performance on syntactic tasks, as measured by BLiMP and BLiMP Supplemental, shows modest variations across different peer counts. For BLiMP, we observe a slight decrease from 1 to 2 peers (51.73\% to 51.55\%), followed by a slight increase with 4 peers (51.58\%). BLiMP Supplemental shows more variation, starting at 53.03\%, dropping to 50.91\% with two peers, and then increasing to 52.25\% with four peers. The average performance across these metrics shows a similar pattern, starting at 51.58\% with one peer, decreasing to 50.62\% with two peers, and slightly recovering to 51.37\% with four peers. These results indicate that while increasing the number of peers does affect performance, the differences are relatively small, with no clear advantage for any particular peer configuration. This suggests that adding more peers may not necessarily lead to substantial gains in syntactic understanding tasks.

\paragraph{Effect of Varying Alpha}
Figure \ref{fig:ablation_study}(b) demonstrates the impact of varying the alpha parameter, which balances the trade-off between cross-entropy loss and peer knowledge distillation in our loss function (Equation \ref{loss}). With $\alpha=0.3$, indicating stronger emphasis on label supervision, we observe the lowest performance. At $\alpha=0.5$, representing an equal balance between label supervision and peer knowledge, performance improves across all metrics. However, when $\alpha=0.7$, shifting focus more towards peer knowledge, we see mixed results with a notable decline in BLiMP (48.65\%) while BLiMP Supplemental shows improvement (52.87\%). This pattern suggests that $\alpha=0.5$ provides an optimal balance: when $\alpha$ is too low (0.3), the models don't fully leverage peer knowledge, and when too high (0.7), excessive reliance on peer learning may compromise individual model performance. The results empirically validate our choice of $\alpha=0.5$ as a balanced configuration for our DWML framework.

\paragraph{Effect of Dynamic Relative Student Importance}
Figure \ref{fig:ablation_study}(c) reveals a positive correlation between dynamically learned importance weights and model performance ($R = 0.7$). Models with higher importance weights demonstrate better accuracy, as shown by the 60M parameter model achieving 50.89\% accuracy with a 0.2511 weight, compared to the 28M model's 49.63\% accuracy with a 0.2484 weight. This near-perfect linear relationship between assigned weights and performance validates our bi-level optimization approach, confirming that the framework successfully identifies and assigns higher weights to more capable models.

\paragraph{Effect of Model Size}
Figure \ref{fig:ablation_study}(c) shows that model performance generally increases with model size, with the 60M parameter model achieving 50.89\% accuracy, followed by 50.13\% for 42M, 49.65\% for 34M, and 49.63\% for 28M parameters. This positive correlation between model size and performance aligns with previous findings, including those from the Chinchilla study \cite{hoffmann_training_2024}. 

\section{Conclusion}
In this paper, we introduced Diversity Induced Weighted Mutual Learning (DWML) as an alternative to teacher-supervised knowledge distillation. While our approach showed modest improvements over the RoBERTa-base baseline, it was the simpler Self-Distillation method that achieved the strongest performance. Our ablation studies on our approach (DWML) revealed that two-peer configurations offered optimal efficiency, a balanced loss function ($\alpha=0.5$) was crucial, and model performance correlated strongly with both dynamically learned importance weights and model size. Regarding computational efficiency, while DWML showed the lowest average GPU utilization, it required longer training times. Hence, in answering our research question about whether student knowledge sharing can match teacher-guided distillation on small datasets, we found that teacher-less methods can indeed match or exceed teacher-supervised approaches, but not necessarily through complex peer learning mechanisms. The success of simpler methods like SD suggests that the field might benefit from focusing on refined single-model approaches rather than elaborate multi-model frameworks. Future work should investigate why simpler teacher-less methods outperform more complex peer learning approaches, explore better neural architecture search techniques, and develop methods to reduce training time while maintaining low resource utilization.

\bibliography{anthology,custom}

\clearpage
\onecolumn
\appendix

\section{Pretraining Hyperparameters}
\begin{table*}[htbp]
\centering
\resizebox{\textwidth}{!}{
\begin{tabular}{@{}lcccccccc@{}}
\toprule
\multirow{2}{*}{Hyperparameters} & \multirow{2}{*}{Base} & \multicolumn{4}{c}{4 peer models} & \multicolumn{2}{c}{2 peer models} & \multirow{2}{*}{1 peer model} \\
\cmidrule(lr){3-6} \cmidrule(lr){7-8}
 & & 1\textsuperscript{*} & 2 & 3\textsuperscript{**} & 4 & 1 & 2 & \\
\midrule
Number of parameters & 125M & 60M & 42M & 34M & 28M & 60M & 42M & 60M \\
Number of layers & 12 & 8 & 16 & 32 & 8 & 8 & 16 & 8 \\
Hidden size & 768 & 512 & 256 & 128 & 256 & 512 & 256 & 512 \\
FF intermediate size & 3072 & 3072 & 3072 & 3072 & 3072 & 3072 & 3072 & 3072 \\
Vocabulary size & 50265 & 50265 & 50265 & 50265 & 50265 & 50265 & 50265 & 50265 \\
Attention heads & 12 & 32 & 8 & 4 & 8 & 32 & 8 & 32 \\
Hidden dropout & 0.1 & 0.1 & 0.1 & 0.1 & 0.1 & 0.1 & 0.1 & 0.1 \\
Attention dropout & 0.1 & 0.1 & 0.1 & 0.1 & 0.1 & 0.1 & 0.1 & 0.1 \\
Training steps & 150 & 150 & 150 & 150 & 150 & 150 & 150 & 150 \\
Mini batch size & 3 & 3 & 3 & 3 & 3 & 3 & 3 & 3 \\
Num. of mini batches & 60 & 60 & 60 & 60 & 60 & 60 & 60 & 60 \\
Sequence length & 514 & 514 & 514 & 514 & 514 & 514 & 514 & 514 \\
Warmup ratio & 0.03\% & 0.03\% & 0.03\% & 0.03\% & 0.03\% & 0.03\% & 0.03\% & 0.03\% \\
Initial learning rate & 0.001 & 0.001 & 0.001 & 0.001 & 0.001 & 0.001 & 0.001 & 0.001 \\
Final learning rate & 0.0001 & 0.0001 & 0.0001 & 0.0001 & 0.0001 & 0.0001 & 0.0001 & 0.0001 \\
Learning rate scheduler & cosine & cosine & cosine & cosine & cosine & cosine & cosine & cosine \\
Weight decay & 0.1 & 0.1 & 0.1 & 0.1 & 0.1 & 0.1 & 0.1 & 0.1 \\
Layer norm $\epsilon$ & 1.00E-12 & 1.00E-12 & 1.00E-12 & 1.00E-12 & 1.00E-12 & 1.00E-12 & 1.00E-12 & 1.00E-12 \\
Optimizer & AdamW & AdamW & AdamW & AdamW & AdamW & AdamW & AdamW & AdamW \\
$\beta_1$ & 0.9 & 0.9 & 0.9 & 0.9 & 0.9 & 0.9 & 0.9 & 0.9 \\
$\beta_2$ & 0.95 & 0.95 & 0.95 & 0.95 & 0.95 & 0.95 & 0.95 & 0.95 \\
Gradient clipping & 1 & 1 & 1 & 1 & 1 & 1 & 1 & 1 \\
\bottomrule
\multicolumn{9}{@{}l@{}}{\textsuperscript{*}Selected for 10M dataset. \textsuperscript{**}Selected for 100M dataset.}
\end{tabular}
}
\caption{Pre-training hyperparameters for base and 4, 2 and 1 peer models for the DWML framework. The same set of hyperparameters are used for other distillation methods for an apple-to-apple comparison.}
\label{tab:pretraining-hyperparameters}
\end{table*}

\section{Finetuning Hyperparameters}
\begin{table*}[htbp]
\centering
\tiny
\begin{tabular}{lc}
\hline
\textbf{Hyperparameters} & \textbf{Full fine-tuning} \\
\hline
Random seed & 643 \\
Batch size & 32 \\
Number of epochs & 6 \\
Dropout & 0.1 \\
Peak learning rate & 2.50E-06 \\
Learning rate decay & cosine \\
Weight decay & 0.1 \\
Optimizer & AdamW \\
Adam $\beta_1$ & 0.9 \\
Adam $\beta_2$ & 0.999 \\
Warmup steps & 3 \\
\hline
\end{tabular}
\caption{Hyperparameters for full fine-tuning the GLUE, SuperGLUE task. We use the same fine-tuning script for comparison of RoBERTa-base and our DWML models. }
\label{tab:finetuning-hyperparameters}
\end{table*}

\newpage
\section{Dataset}
\begin{table*}[htbp]
\centering
\resizebox{\textwidth}{!}{%
\begin{tabular}{@{}llrrr@{}}
\toprule
& & \multicolumn{2}{c}{\textbf{\# Words}} & \\\cmidrule(lr){3-4}
Dataset & Domain & \textsc{strict-small} & \textsc{strict} & Proportion \\
\midrule
CHILDES \citet{macwhinney_childes_2000} & Child-directed speech & 0.44M & 4.21M & 5\% \\
British National Corpus (BNC),\textsuperscript{1} dialogue portion & Dialogue & 0.86M & 8.16M & 8\% \\
Children's Book Test \citet{hill2015goldilocks} & Children's books & 0.57M & 5.55M & 6\% \\
Children's Stories Text Corpus\textsuperscript{2} & Children's books & 0.34M & 3.22M & 3\% \\
Standardized Project Gutenberg Corpus \citet{gerlach2020standardized} & Written English & 0.99M & 9.46M & 10\% \\
OpenSubtitles \citet{creutz2018open} & Movie subtitles & 3.09M & 31.28M & 31\% \\
QCRI Educational Domain Corpus (QED; \citealp{abdelali-etal-2014-amara}) & Educational video subtitles & 1.04M & 10.24M & 11\% \\
Wikipedia\textsuperscript{3} & Wikipedia (English) & 0.99M & 10.08M & 10\% \\
Simple Wikipedia\textsuperscript{4} & Wikipedia (Simple English) & 1.52M & 14.66M & 15\% \\
Switchboard Dialog Act Corpus \citep{stolcke-etal-2000-dialogue} & Dialogue & 0.12M & 1.18M & 1\% \\
\midrule
\emph{Total} & -- & 9.96M & 98.04M & 100\% \\
\bottomrule
\end{tabular}}
\caption{The contents of datasets for the the \textsc{10M} and \textsc{100M} tracks; the table is taken from \citet{}. \textsuperscript{1}\footnotesize\url{http://www.natcorp.ox.ac.uk}\ \ \ \textsuperscript{2}\url{https://www.kaggle.com/datasets/edenbd/children-stories-text-corpus}\ \ \ \textsuperscript{3}\url{https://dumps.wikimedia.org/enwiki/20221220/}\ \ \ \textsuperscript{4}\url{https://dumps.wikimedia.org/simplewiki/20221201/}}
\label{tab:data}
\end{table*}

\section{Detailed results}\label{detailed results}

\subsection{BLiMP}
\begin{table*}[htbp]
\centering
\tiny
\begin{tabular}{lccccccccccccc}
\hline
Method & AA & AS & B & CR & DNA & E & FG & IF & IFS & NPI & Q & SVA & AVG \\
\hline
RoBERTa\_KD\_DWML\_peer1 & 39.60 & 50.40 & 48.70 & 53.70 & 48.50 & 45.10 & 39.10 & 38.10 & 46.90 & 44.40 & 46.80 & 51.10 & 46.00 \\
RoBERTa\_KD\_DWML\_peer2 & 39.50 & 50.30 & 52.60 & 53.70 & 48.40 & 49.40 & 36.70 & 33.20 & 47.10 & 45.50 & 46.30 & 51.20 & 46.20 \\
RoBERTa\_KD\_DWML\_peer3 & 39.40 & 50.30 & 53.60 & 53.70 & 48.30 & 51.40 & 36.70 & 33.30 & 47.80 & 45.10 & 46.30 & 51.30 & 46.40 \\
RoBERTa\_KD\_DWML\_peer4 & 39.80 & 50.60 & 50.10 & 53.10 & 48.70 & 43.60 & 37.30 & 27.50 & 48.20 & 44.30 & 46.10 & 49.90 & 44.90 \\
RoBERTa\_KD\_peer1 & 39.60 & 49.50 & 50.30 & 51.90 & 49.40 & 46.80 & 37.10 & 40.10 & 46.00 & 40.30 & 48.00 & 50.80 & 45.80 \\
RoBERTa\_KD\_peer2 & 39.60 & 50.30 & 53.30 & \textbf{53.90} & 48.50 & 48.60 & 36.50 & 32.70 & 47.20 & 44.70 & 45.90 & 51.20 & 46.10 \\
RoBERTa\_KD\_peer3 & 39.70 & 50.30 & 51.80 & 53.30 & 48.40 & 49.10 & 36.60 & 33.20 & 47.60 & 44.20 & 46.70 & 51.10 & 46.00 \\
RoBERTa\_KD\_peer4 & 54.30 & 49.90 & 52.30 & 56.70 & 45.80 & 48.70 & 37.70 & 32.20 & 48.90 & 44.10 & 49.50 & 48.40 & 47.40 \\
RoBERTa\_SD\_peer1 & \textbf{59.10} & \textbf{51.90} & 48.60 & 47.40 & \textbf{54.10} & 56.90 & 36.50 & \textbf{53.00} & 48.00 & \textbf{66.20} & \textbf{60.70} & 51.20 & \textbf{52.80} \\
RoBERTa\_SD\_peer2 & 45.70 & 53.60 & 58.10 & 53.00 & 51.00 & 52.20 & 37.00 & 52.30 & 47.60 & 53.00 & 38.20 & 50.40 & 50.20 \\
RoBERTa\_SD\_peer3 & 53.50 & 50.10 & 50.50 & 52.30 & 50.50 & 51.70 & \textbf{61.80} & 33.70 & \textbf{57.10} & 42.10 & 36.70 & 49.70 & 49.10 \\
RoBERTa\_SD\_peer4 & 59.10 & 51.40 & 47.70 & 44.90 & 48.30 & 53.10 & 51.30 & 55.70 & 58.30 & 49.40 & 54.80 & 48.80 & 51.90 \\
RoBERTa\_base & 38.90 & 47.90 & \textbf{62.80} & 49.70 & 48.60 & 48.40 & 27.50 & 53.40 & 55.00 & 49.90 & 60.40 & 51.40 & 49.50 \\
DWML\_2model\_peer1 & 45.30 & 51.70 & 57.90 & 48.90 & 47.50 & 50.00 & 46.70 & 45.70 & 58.80 & 43.90 & 55.10 & 50.70 & 50.20 \\
DWML\_2model\_peer2 & 45.30 & 51.70 & 57.70 & 48.90 & 47.50 & 50.10 & 46.60 & 45.70 & 59.50 & 43.80 & 54.90 & 50.70 & 50.20 \\
DWML\_4model\_peer1 & 53.70 & 51.80 & 42.50 & 50.40 & 50.00 & 49.30 & 45.30 & 53.70 & 50.40 & 45.70 & 56.90 & 50.50 & 50.00 \\
DWML\_4model\_peer2 & 53.90 & 51.80 & 42.70 & 50.60 & 50.00 & 49.80 & 45.30 & 53.60 & 50.60 & 50.40 & 57.10 & 50.60 & 50.50 \\
DWML\_4model\_peer3 & 53.60 & 51.70 & 42.00 & 50.60 & 50.00 & 49.70 & 45.20 & 53.50 & 50.60 & 45.40 & 57.10 & 50.60 & 50.00 \\
DWML\_4model\_peer4 & 53.80 & 51.60 & 42.50 & 50.30 & 50.00 & 49.80 & 45.20 & 53.60 & 50.10 & 50.90 & 57.20 & 50.50 & 50.50 \\
DWML\_alpha\_3peer1 & 49.20 & 50.40 & 48.50 & 49.80 & 50.60 & 50.00 & 53.20 & 51.60 & 50.00 & 64.20 & 44.70 & 51.80 & 51.20 \\
DWML\_alpha\_3peer2 & 48.90 & 50.60 & 47.90 & 49.70 & 50.40 & 50.40 & 53.10 & 52.10 & 50.10 & 64.00 & 44.50 & 51.70 & 51.10 \\
DWML\_alpha\_3peer3 & 49.30 & 50.50 & 49.60 & 50.00 & 50.60 & 49.90 & 53.00 & 52.10 & 49.50 & 58.30 & 44.50 & 51.60 & 50.70 \\
DWML\_alpha\_3peer4 & 49.20 & 50.30 & 48.20 & 49.80 & 50.60 & 50.00 & 53.30 & 51.50 & 49.60 & 63.50 & 44.70 & \textbf{51.90} & 51.00 \\
DWML\_alpha\_7peer1 & 58.30 & 49.00 & 40.50 & 49.30 & 52.70 & 54.60 & 50.40 & 56.80 & 43.20 & 41.30 & 54.50 & 49.60 & 50.00 \\
DWML\_alpha\_7peer2 & 58.60 & 49.20 & 40.60 & 49.80 & 52.70 & 54.40 & 50.40 & 57.10 & 43.80 & 41.90 & 57.70 & 49.80 & 50.50 \\
DWML\_alpha\_7peer3 & 58.40 & 49.00 & 39.80 & 49.10 & 52.80 & 54.50 & 50.20 & 56.80 & 44.00 & 41.90 & 54.30 & 49.70 & 50.00 \\
DWML\_alpha\_7peer4 & 58.40 & 49.10 & 40.00 & 49.20 & 52.70 & \textbf{54.80} & 50.20 & 56.80 & 43.90 & 42.30 & 58.60 & 49.70 & 50.50 \\
DML\_peer1 & 54.10 & 49.20 & 52.00 & 50.30 & 48.30 & 47.00 & 42.40 & 47.10 & 54.00 & 27.20 & 48.10 & 49.40 & 47.40 \\
DML\_peer2 & 53.90 & 49.20 & 54.90 & 50.30 & 48.40 & 46.40 & 42.60 & 46.60 & 53.90 & 32.10 & 48.00 & 49.20 & 48.00 \\
DML\_peer3 & 54.00 & 49.10 & 54.70 & 50.60 & 48.40 & 46.90 & 42.50 & 46.70 & 53.80 & 26.90 & 48.00 & 49.00 & 47.60 \\
DML\_peer4 & 54.10 & 49.10 & 54.60 & 50.30 & 48.30 & 46.70 & 42.60 & 46.70 & 53.70 & 26.60 & 48.10 & 49.30 & 47.50 \\
\hline
\label{tab:blimp}
\end{tabular}
\caption{BLiMP results for models trained using different methods. The \textbf{bold} results represent the best model for each task. The metric used is accuracy (\%). Acronyms: AA (Anaphor Agreement), AS (Argument Structure), B (Binding), CR (Control/Raising), DNA (Determiner-Noun Agreement), E (Ellipsis), FG (Filler-Gap), IF (Irregular Forms), IFS (Island Effects), NPI (NPI Licensing), Q (Quantifiers), SVA (Subject-verb agreement)}
\end{table*}

\newpage
\subsection{BLiMP Supplement}
\begin{table*}[htbp]
\centering
\small
\begin{tabular}{lcccccr}
\hline
\textbf{Method} & \textbf{subject\_aux\_} & \textbf{qa\_congruence\_} & \textbf{turn\_} & \textbf{hypernym} & \textbf{qa\_congruence\_} & \textbf{average} \\
 & \textbf{inversion} & \textbf{tricky} & \textbf{taking} & & \textbf{easy} & \\
\hline
KD\_DWML\_peer1 & 44.53 & 60.61 & 56.07 & 52.97 & 46.88 & 52.21 \\
KD\_DWML\_peer2 & 50.61 & 58.79 & 56.07 & 53.09 & 46.88 & 53.09 \\
KD\_DWML\_peer3 & 50.32 & 58.79 & 56.07 & 54.63 & 46.88 & 53.34 \\
KD\_DWML\_peer4 & 49.96 & 58.18 & 55.71 & 52.85 & 51.56 & 53.65 \\
KD\_peer1 & 53.19 & 59.39 & 55.36 & 52.26 & 48.44 & 53.73 \\
KD\_peer2 & 48.00 & 58.18 & 55.71 & 52.85 & 48.44 & 52.64 \\
KD\_peer3 & 48.69 & 59.39 & 56.07 & 53.44 & 45.31 & 52.58 \\
KD\_peer4 & 55.50 & 62.42 & 55.36 & 54.28 & 51.56 & 55.82 \\
SD\_peer1 & 65.48 & 47.88 & 45.00 & \textbf{56.77} & 50.00 & 53.03 \\
SD\_peer2 & 54.20 & 47.88 & 51.79 & 52.85 & \textbf{67.19} & 54.78 \\
SD\_peer3 & 58.81 & 52.12 & 50.71 & 53.68 & 32.81 & 49.63 \\
SD\_peer4 & 66.12 & 59.39 & 52.86 & 54.28 & 50.00 & \textbf{56.53} \\
DWML\_2peer\_1 & 42.40 & \textbf{65.50} & 45.40 & 49.90 & 54.70 & 51.50 \\
DWML\_2peer\_2 & 42.00 & \textbf{65.50} & 46.40 & 50.40 & 51.60 & 51.20 \\
DWML\_alpha\_3peer\_1 & 63.00 & 50.30 & 44.30 & 49.40 & 45.30 & 50.50 \\
DWML\_alpha\_3peer\_2 & 63.10 & 50.90 & 46.10 & 48.80 & 43.80 & 50.50 \\
DWML\_alpha\_3peer\_3 & 60.00 & 50.30 & 44.60 & 50.10 & 42.20 & 49.40 \\
DWML\_alpha\_3peer\_4 & 62.90 & 50.30 & 45.00 & 50.70 & 43.80 & 50.50 \\
DWML\_alpha\_7peer\_1 & 69.70 & 50.90 & 57.50 & 49.60 & 31.30 & 51.80 \\
DWML\_alpha\_7peer\_2 & 70.30 & 52.70 & 57.50 & 51.10 & 32.80 & 52.90 \\
DWML\_alpha\_7peer\_3 & \textbf{70.90} & 50.30 & 57.90 & 51.00 & 31.30 & 52.20 \\
DWML\_alpha\_7peer\_4 & 70.50 & 52.10 & \textbf{59.30} & 51.20 & 32.80 & 53.20 \\
RoBERTa\_base & 54.00 & 41.20 & 52.90 & 51.30 & 45.30 & 48.90 \\
DML\_peer\_1 & 42.60 & 55.80 & 51.40 & 48.90 & 25.00 & 44.70 \\
DML\_peer\_2 & 45.30 & 55.80 & 51.10 & 48.60 & 25.00 & 45.10 \\
DML\_peer\_3 & 42.10 & 57.00 & 51.40 & 50.50 & 25.00 & 45.20 \\
DML\_peer\_4 & 43.80 & 55.20 & 51.40 & 49.40 & 25.00 & 45.00 \\
DWML\_4peer\_1 & 53.6 & 53.4 & 43.2 & 54.4 & 56.6 & 52.25 \\
DWML\_4peer\_2 & 50.6 & 49.8 & 40.2 & 51.1 & 53.3 & 48.99 \\
DWML\_4peer\_3 & 50.6 & 48.4 & 39.5 & 50.4 & 53.1 & 48.43 \\
DWML\_4peer\_4 & 48.8 & 48.0 & 40.0 & 49.6 & 53.6 & 47.99\\
\hline
\end{tabular}
\caption{Supplement BLiMP results for RoBERTa models trained using different distillation methods. All values are presented as percentages.The \textbf{bold} results represent the best model for each task.}
\label{tab:blimp_supp}
\end{table*}

\newpage
\subsection{EWoK}
\begin{table*}[htbp]
\centering
\tiny
\begin{tabular}{lccccccccccc}
\hline
Method & SP & QP & PR & SI & PI & MP & MD & PD & AP & SR & AVG \\
\hline
RoBERTa\_base & 53.5 & 54.5 & 51.0 & 48.0 & 51.8 & 53.5 & 50.1 & 54.2 & 49.5 & 50.4 & 51.6 \\
KD\_peer\_1 & 60.2 & 58.0 & 51.8 & 45.6 & 45.5 & 48.2 & \textbf{52.7} & 32.5 & 45.9 & 50.2 & 50.1 \\
KD\_peer\_2 & 50.4 & 51.9 & 48.9 & 50.7 & 50.2 & 55.3 & 49.7 & 47.5 & 52.6 & 49.2 & 50.3 \\
KD\_peer\_3 & 50.4 & 48.4 & 51.3 & 52.0 & 50.9 & 55.9 & 50.4 & 55.8 & 51.0 & 48.8 & 51.6 \\
KD\_peer\_4 & 38.8 & 43.9 & 50.0 & \textbf{60.5} & 43.7 & 51.2 & 50.8 & 50.0 & 52.6 & 49.4 & 50.4 \\
KD\_DWML\_peer\_1 & \textbf{63.3} & \textbf{62.4} & \textbf{53.8} & 48.3 & 52.5 & \textbf{68.2} & 47.1 & \textbf{61.7} & \textbf{53.8} & 51.0 & \textbf{55.4} \\
KD\_DWML\_peer\_2 & 50.0 & 45.5 & 42.3 & 40.1 & 41.2 & 24.7 & 47.9 & 28.3 & 44.4 & 46.9 & 40.4 \\
KD\_DWML\_peer\_3 & 49.8 & 53.2 & 50.4 & 53.7 & 50.4 & 55.9 & 48.8 & 50.8 & 48.2 & 49.0 & 50.8 \\
KD\_DWML\_peer\_4 & 40.4 & 49.4 & 44.5 & 45.2 & \textbf{54.1} & 57.6 & 49.0 & 40.8 & \textbf{52.0} & 52.0 & 49.8 \\
SD\_peer\_1 & 52.4 & 46.2 & 50.2 & 48.3 & 48.2 & 44.1 & 49.7 & 50.0 & 49.4 & 50.5 & 48.4 \\
SD\_peer\_2 & 52.4 & 47.5 & 49.3 & 47.3 & 51.6 & 45.9 & 50.1 & 54.2 & 49.6 & 50.2 & 49.4 \\
SD\_peer\_3 & 52.0 & 49.4 & 51.2 & 48.3 & 51.1 & 52.4 & 49.0 & 43.3 & 50.4 & 50.9 & 50.4 \\
SD\_peer\_4 & 53.7 & 50.3 & 49.6 & 47.3 & 49.1 & 51.2 & 48.8 & 42.5 & 50.6 & 49.5 & 49.2 \\
DWML\_2peer\_1 & 49.8 & 47.5 & 51.6 & 47.6 & 49.6 & 45.9 & 50.3 & 48.3 & 51.0 & 51.0 & 49.4 \\
DWML\_2peer\_2 & 50.2 & 50.6 & 49.1 & 46.9 & 53.1 & 50.6 & 51.3 & 45.0 & 51.0 & 49.9 & 49.9 \\
DWML\_alpha\_3peer\_1 & 53.5 & 47.8 & 49.0 & 49.0 & 50.4 & 48.8 & 49.6 & 50.0 & 49.4 & 50.3 & 49.5 \\
DWML\_alpha\_3peer\_2 & 51.8 & 49.0 & 49.8 & 46.9 & 50.4 & 50.6 & 49.2 & 49.2 & 50.0 & 50.1 & 49.7 \\
DWML\_alpha\_3peer\_3 & 50.6 & 48.4 & 50.2 & 47.3 & 48.7 & 51.8 & 50.5 & 44.2 & 49.2 & 50.2 & 49.2 \\
DWML\_alpha\_3peer\_4 & 51.4 & 47.5 & 49.9 & 46.6 & 50.4 & 51.2 & 50.8 & 48.3 & 49.9 & 49.8 & 49.4 \\
DWML\_alpha\_7peer\_1 & 50.4 & 54.1 & 50.1 & 50.7 & 49.1 & 55.9 & 50.6 & 49.2 & 50.9 & 51.0 & 51.4 \\
DWML\_alpha\_7peer\_2 & 51.0 & 49.4 & 49.3 & 51.4 & 50.7 & 51.2 & 50.3 & 48.3 & 50.3 & 49.2 & 50.3 \\
DWML\_alpha\_7peer\_3 & 50.0 & 48.4 & 49.3 & 49.0 & 51.6 & 45.3 & 48.4 & 50.0 & 50.1 & 49.9 & 49.6 \\
DWML\_alpha\_7peer\_4 & 50.6 & 49.0 & 49.9 & 49.3 & 50.2 & 51.2 & 48.8 & 49.2 & 49.0 & 49.9 & 49.6 \\
DML\_peer\_1 & 51.4 & 47.1 & 50.0 & 50.0 & 50.5 & 55.9 & 49.1 & 53.3 & 49.2 & 51.0 & 50.1 \\
DML\_peer\_2 & 51.0 & 49.0 & 49.1 & 49.0 & 50.7 & 54.7 & 49.2 & 52.5 & 49.3 & 50.6 & 50.1 \\
DML\_peer\_3 & 49.8 & 51.0 & 49.1 & 52.7 & 46.9 & 52.9 & 51.6 & 44.2 & 49.5 & 50.8 & 50.1 \\
DML\_peer\_4 & 52.7 & 52.5 & 48.4 & 50.3 & 46.8 & 42.4 & 50.6 & 44.2 & 50.0 & 49.7 & 48.8 \\
DWML\_4model\_peer\_1 & 52.0 & 49.7 & 50.0 & 47.6 & 50.9 & 48.8 & 49.5 & 50.0 & 50.1 & 50.3 & 50.0 \\
DWML\_4model\_peer\_2 & 51.2 & 48.7 & 49.0 & 51.4 & 48.9 & 50.0 & 50.6 & 48.3 & 49.0 & 49.9 & 49.8 \\
DWML\_4model\_peer\_3 & 49.4 & 51.3 & 50.4 & 49.7 & 49.5 & 50.6 & 50.1 & 39.2 & 50.0 & 49.6 & 49.1 \\
DWML\_4model\_peer\_4 & 53.9 & 48.7 & 46.8 & 45.2 & 46.8 & 50.6 & 51.4 & 56.7 & 49.7 & 49.2 & 50.3 \\
\hline
\label{tab:ewok}
\end{tabular}
\caption{EWOK evaluation results for different distillation methods. The \textbf{bold} results represent the best performance for each metric. Acronyms: SI (Social Interactions), SP (Social Properties), SR (Social Relations), PI (Physical Interactions), PD (Physical Dynamics), PR (Physical Relations), MD (Material Dynamics), MP (Material Properties), AP (Agent Properties), QP (Quantitative Properties). The metric used is accuracy, and results are presented as percentage values.The \textbf{bold} results represent the best model for each task.}
\end{table*}

\newpage
\section{Peer importance training during distillation}
\begin{figure}[hbt!]
    \centering
    \includegraphics[width=0.5\linewidth]{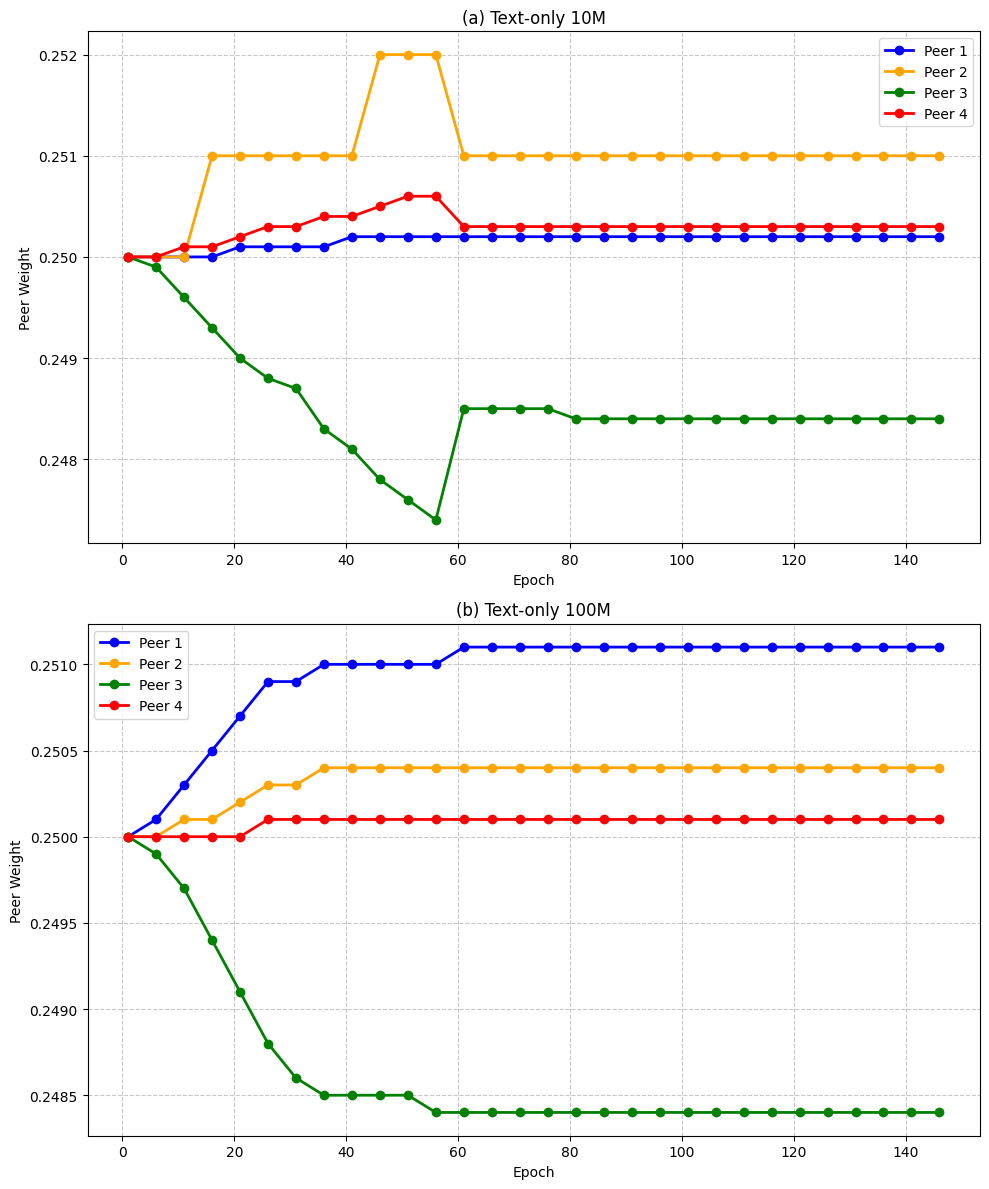}
    \caption{Peer importance weights dynamically trained using mirror descent algorithm as described in Equation \ref{mirror}}
    \label{fig:weights evolution}
\end{figure}

\section{GPU Utilization}

\begin{table}[hbt!]
\centering
\begin{tabular}{lccc}
\hline
\textbf{Method} & \textbf{Avg. GPU util.(\%)}$\downarrow$ & \textbf{Avg. FLOPs/peer(T)}$\downarrow$ \\
\hline
RoBERTa-base & 64.00 & 290.4 \\
KD-DWML & 62.71 (-2.0\%) & 269.6 (-7.2\%) \\
KD & 57.53 (-10.1\%) & 263.7 (-9.2\%) \\
SD & 55.24 (-13.7\%) & 177.75 (-38.8\%) \\
DWML & 43.20 (-32.5\%) & 124.4 (-57.2\%) \\
DML & 43.66 (-31.8\%) & 118.5 (-59.2\%) \\
\hline
\end{tabular}
\caption{Comparison of average GPU utilization (in \%) and FLOPs per peer (in Trillion) across different methods. Percentages in parentheses show reduction relative to RoBERTa-base baseline. Lower values indicate better efficiency.}
\label{tab:training_utilization}
\end{table}

\begin{figure}[hbt!]
    \includegraphics[width=1\linewidth]{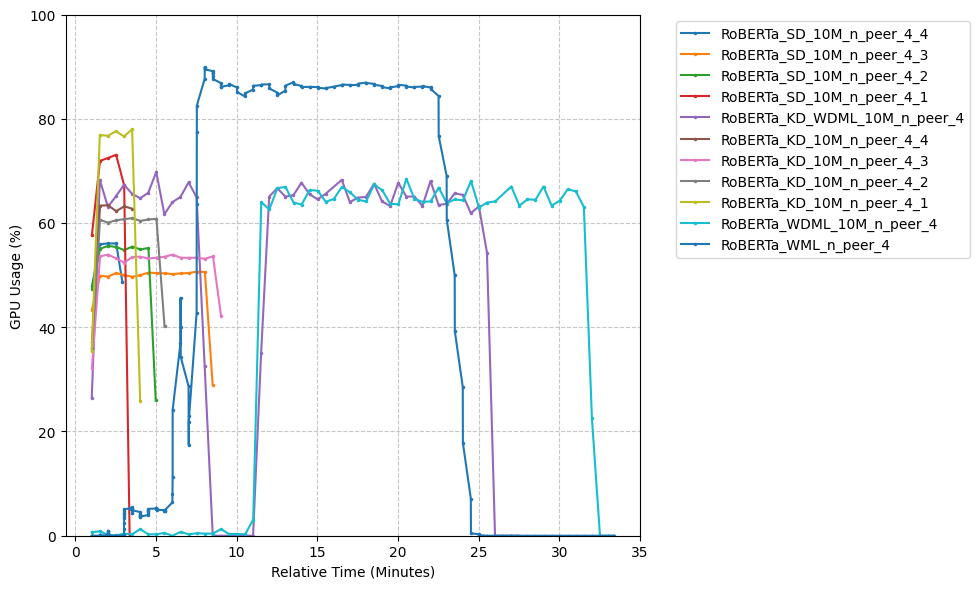}
    \caption{GPU utilization for different distillation methods.}
    \label{fig:gpu_utilization}
\end{figure}

\end{document}